\documentclass{icga}

\usepackage{amssymb}
\usepackage{amsmath}
\usepackage{graphicx}
\usepackage{tikz}
\usepackage{caption}
\usepackage{subcaption}
\usepackage{wrapfig}
\usepackage[pdfusetitle]{hyperref}

\usetikzlibrary{shapes}

\title{ENDGAME ANALYSIS OF DOU SHOU QI}
\runningtitle{Endgame Analysis of Dou Shou Qi}
\author{Jan N. van Rijn\thanks{Leiden Institute of Advanced Computer Science,
Universiteit Leiden, The Netherlands. E-mail: j.n.van.rijn@liacs.leidenuniv.nl} \and
Jonathan K. Vis\thanks{Leiden Institute of Advanced Computer Science,
Universiteit Leiden, The Netherlands. E-mail: j.k.vis@liacs.leidenuniv.nl}}
\affiliation{Leiden, The Netherlands}
\issue{June 2014}
\hypersetup{pdftitle={Endgame Analysis of Dou Shou Qi},pdfauthor={Jan N. van Rijn, Jonathan K. Vis}}

\begin{document}
\maketitle
\setcounter{page}{120}

\begin{abstract}
Dou Shou Qi is a game in which two players control a number of pieces,
each of them aiming to move one of their pieces onto a given square. 
We implemented an engine for analyzing the game. Moreover, we created 
a series of endgame tablebases containing all configurations with 
up to four pieces. These tablebases are the first steps towards 
theoretically solving the game. Finally, we constructed decision 
trees based on the endgame tablebases. In this note we report on 
some interesting patterns.
\end{abstract}

\section{Introduction}\label{sec:intro} 
\emph{Dou Shou Qi} (meaning: ``Game of Fighting Animals'') is a 
Chinese board game first described by~\citeaby{Pritchard2007}. 
In the Western world it is often
called Jungle, The Jungle Game, Jungle Chess, or Animal Chess.
Dou Shou Qi is a two-player abstract strategy game. It contains
some elements from Chess and Stratego as well as some other Chess-like
Chinese games (e.g., Banqi). Its origins are not entirely clear, but
it seems that it evolved rather recently (around the 1900s) in China.
Dou Shou Qi is played on a rectangular board consisting of
$9\times7$~squares (see Figure~\ref{fig:board}). The columns are called
\emph{files} and are labelled $a$--$g$ from left to right. The rows or
\emph{ranks} are numbered $1$--$9$ from bottom to top (the board is
viewed from the position of the white player).

\begin{wrapfigure}{r}{0.5\textwidth}
%\begin{figure}[!ht]
\begin{center}
\begin{tikzpicture}[scale=0.9]
\draw [step=1, very thin] (0, 0) grid (7, 9);
\draw [fill] (2, 8) rectangle (3, 9);
\draw [fill] (4, 8) rectangle (5, 9);
\draw [fill] (3, 7) rectangle (4, 8);
\draw [draw=gray, fill=black] (3, 8) rectangle (4, 9);

\node at (3.5, 0.5) {\textbf{D}};
\node at (2.5, 0.5) {\textbf{T}};
\node at (4.5, 0.5) {\textbf{T}};
\node at (3.5, 1.5) {\textbf{T}};

\node at (1.5, 3.5) {\textbf{W}};
\node at (2.5, 3.5) {\textbf{W}};
\node at (1.5, 4.5) {\textbf{W}};
\node at (2.5, 4.5) {\textbf{W}};
\node at (1.5, 5.5) {\textbf{W}};
\node at (2.5, 5.5) {\textbf{W}};

\node at (4.5, 3.5) {\textbf{W}};
\node at (5.5, 3.5) {\textbf{W}};
\node at (4.5, 4.5) {\textbf{W}};
\node at (5.5, 4.5) {\textbf{W}};
\node at (4.5, 5.5) {\textbf{W}};
\node at (5.5, 5.5) {\textbf{W}};

\node [text=white] at (3.5, 8.5) {\textbf{D}};
\node [text=white] at (2.5, 8.5) {\textbf{T}};
\node [text=white] at (4.5, 8.5) {\textbf{T}};
\node [text=white] at (3.5, 7.5) {\textbf{T}};

\node [circle, draw] at (0.5, 2.5) {\textbf{8}};
\node [circle, draw] at (0.5, 0.5) {\textbf{6}};
\node [circle, draw] at (1.5, 1.5) {\textbf{2}};
\node [circle, draw] at (2.5, 2.5) {\textbf{3}};
\node [circle, draw] at (4.5, 2.5) {\textbf{5}};
\node [circle, draw] at (5.5, 1.5) {\textbf{4}};
\node [circle, draw] at (6.5, 2.5) {\textbf{1}};
\node [circle, draw] at (6.5, 0.5) {\textbf{7}};

\node [circle, fill, text=white] at (0.5, 8.5) {\textbf{7}};
\node [circle, fill, text=white] at (0.5, 6.5) {\textbf{1}};
\node [circle, fill, text=white] at (1.5, 7.5) {\textbf{4}};
\node [circle, fill, text=white] at (2.5, 6.5) {\textbf{5}};
\node [circle, fill, text=white] at (4.5, 6.5) {\textbf{3}};
\node [circle, fill, text=white] at (5.5, 7.5) {\textbf{2}};
\node [circle, fill, text=white] at (6.5, 8.5) {\textbf{6}};
\node [circle, fill, text=white] at (6.5, 6.5) {\textbf{8}};

\end{tikzpicture}
\caption[Dou Shou Qi game board]{Dou Shou Qi game board.}
\label{fig:board}
\end{center}
%\end{figure}
\end{wrapfigure}

There are four different kinds of squares: den, trap, water, and land. 
There are two \emph{dens}~(D) located in the center of the first and the
last rank ($d1$ and $d9$). Each den is surrounded by
\emph{traps}~(T). There are also two rectangular
($3\times2$~squares) bodies of \emph{water}~(W) sometimes
called \emph{rivers}. The remaining squares are ordinary land squares.
Each player has eight different pieces representing different animals (see below).
Each animal has a certain \emph{strength}, according to which they can
\emph{capture} other (opponent's) pieces. Only pieces with the same or
a higher strength may capture an opponent's piece. The only exception
to this rule regards the weakest (rat) and the strongest (elephant)
pieces. Just like the spy in Stratego, the weakest piece may capture
the strongest. The strength of the pieces, from weak to strong, is:
$1$ \texttt{R}, \texttt{r} --- Rat (sometimes called mouse);
$2$ \texttt{C}, \texttt{c} --- Cat;
$3$ \texttt{W}, \texttt{w} --- Wolf (sometimes called fox);
$4$ \texttt{D}, \texttt{d} --- Dog;
$5$ \texttt{P}, \texttt{p} --- Panther (sometimes called leopard);
$6$ \texttt{T}, \texttt{t} --- Tiger;
$7$ \texttt{L}, \texttt{l} --- Lion;
$8$ \texttt{E}, \texttt{e} --- Elephant.
The initial placement of the pieces is fixed, see
Figure~\ref{fig:board}. The capital letters are used to
denote the white pieces. Players alternate moves with White moving
first. Each turn one piece must be moved.
Each piece can move one square either horizontally or vertically. In
principle, a piece may not move into the water, and it is also
forbidden to enter its own den ($d1$ for White, and $d9$ for Black).
The rat is the only piece that can swim, and is therefore able to
enter the water. It may also capture in the water (the opponent's
rat). However, it may not capture the elephant from the water.
Lions and tigers are able to leap over water (either horizontally or
vertically). They cannot jump over the water when a rat (own or
opponent's) is on any of the intermediate water squares.
When a piece is in an opponent's trap ($c9$, $d8$, $e9$ for White and
$c1$, $d2$, $e1$ for Black), its strength is effectively reduced to
zero, meaning that any of the opponent's pieces may capture it
regardless of its strength. A piece in one of its own traps is
unaffected.
The objective of the game is either to place one of the pieces in the
opponent's den or to eliminate all of the opponent's pieces. As in
Chess, stalemate positions are declared a draw. A threefold repetition
rule is imposed in some variants of this game. However, the existence of such a
rule is irrelevant for our analyses.

The game Dou Shou Qi is not extensively studied in the literature.
\citeaby{Burnett2010} attempts 
to characterize certain local properties of subproblems that occur. 
These so-called loosely coupled subproblems
can be analyzed separately in contrast to analyzing the problem as a
whole resulting in a possible speed-up in the overall analysis. The
author also proposes an evaluation (utility) function for Dou Shou Qi,
which we will use in our research as well. We will also present an
engine without taking the loosely coupled subproblems into account.
The game has been proven PSPACE-hard by~\citeaby{Rijn2012} and \citeaby{Hoogeboom2014}. 

\section{Dou Shou Qi Engine}\label{sec:bagheera}

To get a feeling for the search complexity of a Dou Shou Qi game, we
present some numbers. An average configuration allows for $\approx20$~legal
moves (out of a maximum of~$32$). In theory a complete game tree of
all possible games from the initial configuration can be constructed
by recursively applying the rules of the game. From the initial
configuration the number of leaves visited per ply can be found in
Table~\ref{tab:initial_leaves}. Assuming an average game length of
$40$~moves ($80$~plies), there are approximately $20^{80}$ possible
games.

In this section we introduce a Dou Shou Qi (analyzing)
engine\footnote{For the implementation of the engine and the
retrograde analysis see:
\texttt{http://www.liacs.nl/home/jvis/doushouqi}.}, similar
to a Chess engine which is used to search through the game tree given
a certain configuration. The seminal 1950 paper by \citeaby{Shannon1950} 
lists the elements of a Chess-playing
computer (also known as an engine), which are also applicable to a
Dou Shou Qi engine as both games are similar. Usually, an engine
consists of three parts: a move generator, which generates a set of
legal moves given a configuration; an evaluation function (or utility
function), which is able to assign a value to the leaf of the game
tree, and a search algorithm to traverse the game tree. As evaluation
function we use the method constructed by~\citeaby{Burnett2010}.

\begin{table}[ht!]
\small
\begin{center}
\caption{The number of leaf nodes that are evaluated by the minimax method (no pruning) 
         at a certain depth from the initial configuration. The performance 
         was measured on an Intel i7-2600 with $16$~GB~RAM.}
\label{tab:initial_leaves}
\begin{tabular}{r r r r l}
ply &          time &  number of leaf nodes \\
\hline
$1$ &        $0.00$ &                  $24$ \\
$2$ &        $0.00$ &                 $576$ \\
$3$ &        $0.00$ &            $12,\!240$ \\
$4$ &        $0.06$ &           $260,\!100$ \\
$5$ &        $1.26$ &       $5,\!098,\!477$ \\
$6$ &       $23.46$ &      $99,\!860,\!517$ \\
$7$ & $7\!:\!51.33$ & $1,\!890,\!415,\!534$ \\
\end{tabular}
\end{center}
\end{table}

In most Chess-playing engines today some form of the minimax algorithm
is used. These engines try to minimize the possible loss for a worst
case (maximum loss) scenario. As is well-known, the performance of the minimax algorithm
can be improved by the use of alpha-beta pruning. Here, a branch is not
further evaluated when at least one of the immediately following
configurations proves to be worse (in terms of the evaluation
function) than a previously examined move, cf.~\citeaby{Knuth1976}.
In our case, using the same machine, we are able to search the game
tree within the eight minutes to a depth of $14$~plies (instead of $7$~plies by the minimax method).
We remark that the aforementioned search methods operate on trees, while the actual
search space is an acyclic graph. Configurations that have been
considered previously might be considered again by means of so-called
transpositions. A reordered sequence of the same set of moves results
in the same configuration. This is especially true for end game
configurations where a few pieces can move around in many ways to form
equal configurations. By storing evaluated configurations in memory we
can omit expensive re-searches of the same configuration. Commonly, a
hash table is used. \citeaby{Zobrist1990} introduced a hashing
method for Chess which can easily be extended to a more general case.
%The idea is to combine random numbers ($64$-bits) for each piece at
%each location by using the XOR operator. When using a set of carefully
%constructed random numbers the change of two different configurations
%that result in the same hash key can be ignored. Another nice property
%of this method is that a new key (after a move) can be constructed
%from the old key (current configuration). In the transposition table
%we store the hash key together with information about this
%configuration, i.e., the evaluated score (either exact or a bounding
%score in case of alpha-beta pruning), the depth the configuration was
%evaluated for, as shallower results are less trustworthy than deeper
%searches.
%The transposition table size is obviously smaller than all $64$-bit
%possible hash keys. We map the hash key to an entry in the table by
%taking the values of the key modulo the table size as its index.

Our engine consists of the alpha-beta algorithm augmented with 
a (large) transposition table using the
Zobrist hashing method. This engine has proven its usefulness. 
For instance, we constructed endgame tablebases (see
Section~\ref{sec:representation}), which in turn can be used to improve the
engine.

\section{Endgame Tablebase Construction}\label{sec:representation}
An endgame tablebase describes for every configuration with a
certain number of pieces, relevant precalculated information. For this game it states which side has a
theoretical win or that the game is drawn.
Endgame tablebases exist for many well-known games, most notably for
Chess by~\citeaby{Thompson86} and
Checkers by~\citeaby{Schaeffer04}. 
The technique we use for constructing this database is retrograde
analysis, similar to the technique used by~\citeaby{Thompson86}. 

The resulting endgame tablebase contains for each configuration up to
four pieces the game theoretic value, the amount of moves until the
theoretic value has been achieved, and the first move (best move) that
will lead to this result.
Some general statistics about the endgame tablebase are shown in
Table~\ref{tab:tablebase_summary}. Each row summarizes the number of
configurations it contains, and the distribution of obtainable results
for the player to move. Furthermore, the longest sequence of moves
that leads to a forced win for either player is also displayed.

\begin{table}[!ht]
\small
\begin{center}
\caption{Summary of the endgame tablebase up to four pieces.}
\label{tab:tablebase_summary}
\begin{tabular}{r r r r r r}
pieces &             positions &                  wins &                losses &             draws & longest sequence \\
\hline
   $2$ &           $160,\!068$ &            $82,\!852$ &            $64,\!501$ &        $12,\!715$ &             $34$ \\
   $3$ &      $54,\!354,\!684$ &      $30,\!297,\!857$ &      $23,\!369,\!820$ &       $687,\!007$ &             $67$ \\
   $4$ & $9,\!685,\!020,\!510$ & $5,\!468,\!841,\!129$ & $4,\!001,\!236,\!829$ & $214,\!942,\!568$ &            $117$ \\
\end{tabular}
\end{center}
\end{table}

The endgame tablebase can be split into various \emph{partitions}, based 
on which pieces are involved. After extracting \emph{game features} 
from the configurations, a decision tree can be constructed for each 
partition using a technique such as described by~\citeaby{Quinlan86}. %TODO: why game features?
Given a configuration, we can calculate the game features from it and
using the decision tree associated with the pieces in the configuration,
determine the game theoretic value. Some resulting decision trees are
shown in Figure~\ref{fig:decisiontrees}. The features that were used for
constructing these trees are listed in Table~\ref{tab:features}. Storing
such a decision tree in memory takes less space then storing the entire
endgame tablebase. Furthermore, this representation can yield interesting
observations about the game.

\begin{table}[!ht]
\small
\begin{center}
\caption{Game features extracted from the endgame tablebase used as split criteria.}
\label{tab:features}
\begin{tabular}{l l l}
feature & values & description \\
\hline
   closest & $\{\mathrm{white}, \mathrm{black}\}$ & the player that can reach the opposing den first  \\
   unopposed$_\mathrm{\{w,b\}}$ & boolean & the piece~$\{\mathrm{w},\mathrm{b}\}$ can reach the den unopposed \\
   sector$_\mathrm{\{w,b\}}$ & $\{\mathrm{top},\mathrm{mid},\mathrm{bot}\}$ & $\mathrm{top}$: rank $\ge 7$, $\mathrm{mid}$: $3 <$ rank $< 7$, $\mathrm{bot}$: rank $\le 3$;
          rank of piece~$\{\mathrm{w},\mathrm{b}\}$ \\
   distance$_\mathrm{d}$ & integer $[0$--$11]$ & Manhattan distance between the white piece and the white den \\
   distance$_\mathrm{p}$ & integer $[0$--$14]$ & Manhattan distance between the two pieces \\
   parity & $\{0, 1\}$ & distance$_\mathrm{p}$ $\mathrm{mod}\,2$ \\
   adjacent & boolean & distance$_\mathrm{p}$ $= 1$ \\
   trapped & boolean & Black is trapped \\
   can cross & boolean & White can reach rank~$7$ unopposed on the shortest route to the black den \\
\end{tabular}
\end{center}
\end{table}

The feature \emph{closest} determines which player has a piece closest
to the opposing den, and therefore can reach it first. When both
pieces are at the same distance, White is marked as closest 
since he moves first. The feature \emph{unopposed} determines whether the player
can reach the opposing den before the other player can get there. For
White this is true iff there exists a black trap for which the
manhattan distance between the white piece and that trap is smaller
than the manhattan distance between the black piece and that trap
(Black cannot move through its own den). The feature \emph{sector} is
derived directly from the rank of a piece, and divides the board
evenly into three sectors. The feature \emph{can cross} states whether
the white piece can reach rank~$7$ unopposed, in such a way that it
still minimises the number of moves to the opposing den.

\section{Endgame Analysis}\label{sec:analysis}

\begin{figure}[!ht]
  \begin{center}
    %\begin{subfigure}[t]{0.33\textwidth}
      \begin{tikzpicture}[yscale=-0.7, xscale=0.37]
\footnotesize
\node (closest) [ellipse, draw, inner sep=1pt] at (0, 0) {closest};
\node (wunopposed) [ellipse, draw, inner sep=1pt] at (-3, 1.5) {unopposed$_\mathrm{w}$};
\node (bunopposed) [ellipse, draw, inner sep=1pt] at (3, 1.5) {unopposed$_\mathrm{b}$};
\node (parity1) [ellipse, draw, inner sep=1pt] at (-4.2, 3) {parity};
\node (leaf1) [rectangle, draw, inner sep=2pt] at (-1.5, 3) {white};
\node (parity2) [ellipse, draw, inner sep=1pt] at (1.5, 3) {parity};
\node (leaf2) [rectangle, draw, inner sep=2pt] at (4.2, 3) {black};
\node (leaf3) [rectangle, draw, inner sep=2pt] at (-5.2, 4.5) {black};
\node (leaf4) [rectangle, draw, inner sep=2pt] at (-2.8, 4.5) {white};
\node (leaf5) [rectangle, draw, inner sep=2pt] at (0.3, 4.5) {black};
\node (leaf6) [rectangle, draw, inner sep=2pt] at (2.8, 4.5) {white};

\node (phantom) at (0, 10) {
\small (a) Equal material
};

\draw (closest) to node [left] {$\mathrm{white}$} (wunopposed);
\draw (closest) to node [right] {$\mathrm{black}$} (bunopposed);
\draw (wunopposed) to node [left] {false} (parity1);
\draw (wunopposed) to node [right] {true} (leaf1);
\draw (bunopposed) to node [left] {false} (parity2);
\draw (bunopposed) to node [right] {true} (leaf2);
\draw (parity1) to node [left] {0} (leaf3);
\draw (parity1) to node [right] {1} (leaf4);
\draw (parity2) to node [left] {0} (leaf5);
\draw (parity2) to node [right] {1} (leaf6);

\end{tikzpicture}
    %\end{subfigure}
    %\begin{subfigure}[t]{0.33\textwidth}
      \begin{tikzpicture}[yscale=-0.7, xscale=0.37]
\footnotesize
\node (closest) [ellipse, draw, inner sep=1pt] at (0, 0) {closest};
\node (unopposed) [ellipse, draw, inner sep=1pt] at (-2.5, 1.5) {unopposed$_\mathrm{w}$};
\node (adjacent) [ellipse, draw, inner sep=1pt] at (2.5, 1.5) {adjacent};
\node (parity) [ellipse, draw, inner sep=1pt] at (-5, 3) {parity};
\node (leaf1) [rectangle, draw, inner sep=2pt] at (-1.5, 3) {white};
\node (leaf2) [rectangle, draw, inner sep=2pt] at (2, 3) {black};
\node (trapped) [ellipse, draw, inner sep=1pt] at (5, 3) {trapped};
\node (cross) [ellipse, draw, inner sep=1pt] at (-5, 4.5) {can cross};
\node (leaf3) [rectangle, draw, inner sep=2pt] at (-1.5, 4.5) {black};
\node (leaf4) [rectangle, draw, inner sep=2pt] at (3.8, 4.5) {black};
\node (leaf5) [rectangle, draw, inner sep=2pt] at (6.2, 4.5) {white};
\node (pdistance) [ellipse, draw, inner sep=1pt] at (-5, 6) {distance$_\mathrm{p}$};
\node (leaf6) [rectangle, draw, inner sep=2pt] at (-1.5, 6) {draw};
\node (ddistance) [ellipse, draw, inner sep=1pt] at (-5, 7.5) {distance$_\mathrm{d}$};
\node (leaf7) [rectangle, draw, inner sep=2pt] at (-1.5, 7.5) {black};
\node (leaf8) [rectangle, draw, inner sep=2pt] at (-6.1, 9) {black};
\node (leaf9) [rectangle, draw, inner sep=2pt] at (-3.9, 9) {draw};

\node (phantom) at (0, 10) {
\small (b) Black stronger (no rats, lions or tigers)
};

\draw (closest) to node [left] {$\mathrm{white}$} (unopposed);
\draw (closest) to node [right] {$\mathrm{black}$} (adjacent);
\draw (unopposed) to node [left] {false} (parity);
\draw (unopposed) to node [right] {true} (leaf1);
\draw (adjacent) to node [left] {false} (leaf2);
\draw (adjacent) to node [right] {true} (trapped);
\draw (parity) to node [left] {1} (cross);
\draw (parity) to node [right] {0} (leaf3);
\draw (trapped) to node [left] {false} (leaf4);
\draw (trapped) to node [right] {true} (leaf5);
\draw (cross) to node [left] {false} (pdistance);
\draw (cross) to node [right] {true} (leaf6);
\draw (pdistance) to node [left] {$>10$} (ddistance);
\draw (pdistance) to node [right] {$\le10$} (leaf7);
\draw (ddistance) to node [left] {$\le3$} (leaf8);
\draw (ddistance) to node [right] {$>3$} (leaf9);

\end{tikzpicture}
    %\end{subfigure}
    %\begin{subfigure}[t]{0.33\textwidth}
      \begin{tikzpicture}[yscale=-0.7, xscale=0.37]
\footnotesize
\node (closest) [ellipse, draw, inner sep=1pt] at (0, 0) {closest};
\node (unopposed) [ellipse, draw, inner sep=1pt] at (-2.5, 1.5) {unopposed$_\mathrm{w}$};
\node (adjacent) [ellipse, draw, inner sep=1pt] at (2.5, 1.5) {adjacent};
\node (bposition) [ellipse, draw, inner sep=1pt] at (-5, 3) {sector$_\mathrm{b}$};
\node (leaf1) [rectangle, draw, inner sep=2pt] at (-1.5, 3) {white};
\node (leaf2) [rectangle, draw, inner sep=2pt] at (2, 3) {black};
\node (trapped) [ellipse, draw, inner sep=1pt] at (5, 3) {trapped};
\node (wposition) [ellipse, draw, inner sep=1pt] at (-5, 4.5) {sector$_\mathrm{w}$};
\node (leaf3) [rectangle, draw, inner sep=2pt] at (-1.5, 4.5) {draw};
\node (leaf4) [rectangle, draw, inner sep=2pt] at (3.8, 4.5) {black};
\node (leaf5) [rectangle, draw, inner sep=2pt] at (6.2, 4.5) {white};
\node (leaf6) [rectangle, draw, inner sep=2pt] at (-6.1, 6) {black};
\node (leaf7) [rectangle, draw, inner sep=2pt] at (-3.9, 6) {draw};

\node (phantom) at (0, 10) {
\small (c) White lion vs. black elephant
};

\draw (closest) to node [left] {$\mathrm{white}$} (unopposed);
\draw (closest) to node [right] {$\mathrm{black}$} (adjacent);
\draw (unopposed) to node [left] {false} (parity);
\draw (unopposed) to node [right] {true} (leaf1);
\draw (adjacent) to node [left] {false} (leaf2);
\draw (adjacent) to node [right] {true} (trapped);
\draw (bposition) to node [left] {} (wposition);
\draw (bposition) to node [right] {$\mathrm{top}$} (leaf3);
\draw (trapped) to node [left] {false} (leaf4);
\draw (trapped) to node [right] {true} (leaf5);
\draw (wposition) to node [left] {$\mathrm{mid}$} (leaf6);
\draw (wposition) to node [right] {} (leaf7);

\end{tikzpicture}
    %\end{subfigure}
  \vspace{-.5cm}
  \caption{Decision trees for partitions containing two pieces.}
  \label{fig:decisiontrees}
  \end{center}
\end{figure}
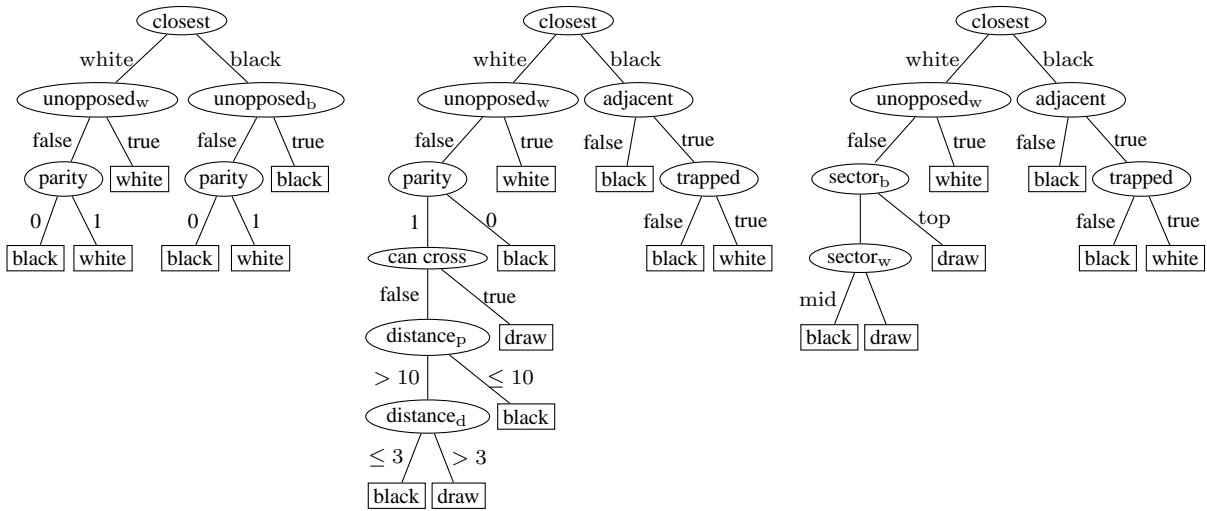

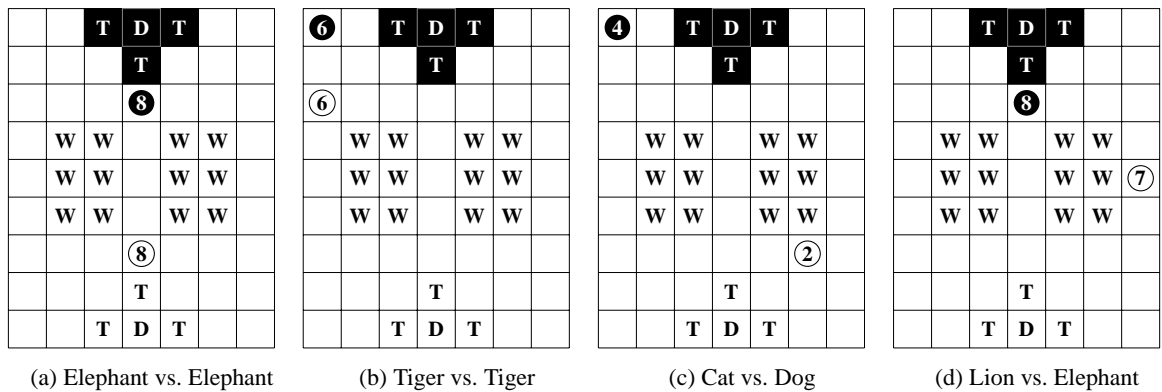
\begin{figure}[!ht]
  \begin{center}
    \begin{subfigure}[b]{0.24\textwidth}
      \begin{tikzpicture}[scale=0.5]
\footnotesize
\draw [step=1, very thin] (0, 0) grid (7, 9);
\draw [fill] (2, 8) rectangle (3, 9);
\draw [fill] (4, 8) rectangle (5, 9);
\draw [fill] (3, 7) rectangle (4, 8);
\draw [draw=gray, fill=black] (3, 8) rectangle (4, 9);

\node at (3.5, 0.5) {\textbf{D}};
\node at (2.5, 0.5) {\textbf{T}};
\node at (4.5, 0.5) {\textbf{T}};
\node at (3.5, 1.5) {\textbf{T}};

\node at (1.5, 3.5) {\textbf{W}};
\node at (2.5, 3.5) {\textbf{W}};
\node at (1.5, 4.5) {\textbf{W}};
\node at (2.5, 4.5) {\textbf{W}};
\node at (1.5, 5.5) {\textbf{W}};
\node at (2.5, 5.5) {\textbf{W}};

\node at (4.5, 3.5) {\textbf{W}};
\node at (5.5, 3.5) {\textbf{W}};
\node at (4.5, 4.5) {\textbf{W}};
\node at (5.5, 4.5) {\textbf{W}};
\node at (4.5, 5.5) {\textbf{W}};
\node at (5.5, 5.5) {\textbf{W}};

\node [text=white] at (3.5, 8.5) {\textbf{D}};
\node [text=white] at (2.5, 8.5) {\textbf{T}};
\node [text=white] at (4.5, 8.5) {\textbf{T}};
\node [text=white] at (3.5, 7.5) {\textbf{T}};

\node [circle, draw, inner sep=1pt] at (3.5, 2.5) {\textbf{8}};

\node [circle, fill, text=white, inner sep=1pt] at (3.5, 6.5) {\textbf{8}};

\end{tikzpicture}
      \caption{Elephant vs. Elephant}
      \label{fig:evse}
    \end{subfigure}
    \begin{subfigure}[b]{0.24\textwidth}
      \begin{tikzpicture}[scale=0.5]
\footnotesize
\draw [step=1, very thin] (0, 0) grid (7, 9);
\draw [fill] (2, 8) rectangle (3, 9);
\draw [fill] (4, 8) rectangle (5, 9);
\draw [fill] (3, 7) rectangle (4, 8);
\draw [draw=gray, fill=black] (3, 8) rectangle (4, 9);

\node at (3.5, 0.5) {\textbf{D}};
\node at (2.5, 0.5) {\textbf{T}};
\node at (4.5, 0.5) {\textbf{T}};
\node at (3.5, 1.5) {\textbf{T}};

\node at (1.5, 3.5) {\textbf{W}};
\node at (2.5, 3.5) {\textbf{W}};
\node at (1.5, 4.5) {\textbf{W}};
\node at (2.5, 4.5) {\textbf{W}};
\node at (1.5, 5.5) {\textbf{W}};
\node at (2.5, 5.5) {\textbf{W}};

\node at (4.5, 3.5) {\textbf{W}};
\node at (5.5, 3.5) {\textbf{W}};
\node at (4.5, 4.5) {\textbf{W}};
\node at (5.5, 4.5) {\textbf{W}};
\node at (4.5, 5.5) {\textbf{W}};
\node at (5.5, 5.5) {\textbf{W}};

\node [text=white] at (3.5, 8.5) {\textbf{D}};
\node [text=white] at (2.5, 8.5) {\textbf{T}};
\node [text=white] at (4.5, 8.5) {\textbf{T}};
\node [text=white] at (3.5, 7.5) {\textbf{T}};

\node [circle, draw, inner sep=1pt] at (0.5, 6.5) {\textbf{6}};

\node [circle, fill, text=white, inner sep=1pt] at (0.5, 8.5) {\textbf{6}};

\end{tikzpicture}
      \caption{Tiger vs. Tiger}
      \label{fig:tvst}
    \end{subfigure}
    \begin{subfigure}[b]{0.24\textwidth}
      \begin{tikzpicture}[scale=0.5]
\footnotesize
\draw [step=1, very thin] (0, 0) grid (7, 9);
\draw [fill] (2, 8) rectangle (3, 9);
\draw [fill] (4, 8) rectangle (5, 9);
\draw [fill] (3, 7) rectangle (4, 8);
\draw [draw=gray, fill=black] (3, 8) rectangle (4, 9);

\node at (3.5, 0.5) {\textbf{D}};
\node at (2.5, 0.5) {\textbf{T}};
\node at (4.5, 0.5) {\textbf{T}};
\node at (3.5, 1.5) {\textbf{T}};

\node at (1.5, 3.5) {\textbf{W}};
\node at (2.5, 3.5) {\textbf{W}};
\node at (1.5, 4.5) {\textbf{W}};
\node at (2.5, 4.5) {\textbf{W}};
\node at (1.5, 5.5) {\textbf{W}};
\node at (2.5, 5.5) {\textbf{W}};

\node at (4.5, 3.5) {\textbf{W}};
\node at (5.5, 3.5) {\textbf{W}};
\node at (4.5, 4.5) {\textbf{W}};
\node at (5.5, 4.5) {\textbf{W}};
\node at (4.5, 5.5) {\textbf{W}};
\node at (5.5, 5.5) {\textbf{W}};

\node [text=white] at (3.5, 8.5) {\textbf{D}};
\node [text=white] at (2.5, 8.5) {\textbf{T}};
\node [text=white] at (4.5, 8.5) {\textbf{T}};
\node [text=white] at (3.5, 7.5) {\textbf{T}};

\node [circle, draw, inner sep=1pt] at (5.5, 2.5) {\textbf{2}};

\node [circle, fill, text=white, inner sep=1pt] at (0.5, 8.5) {\textbf{4}};

\end{tikzpicture}
      \caption{Cat vs. Dog}
      \label{fig:cvsd}
    \end{subfigure}
    \begin{subfigure}[b]{0.24\textwidth}
      \begin{tikzpicture}[scale=0.5]
\footnotesize
\draw [step=1, very thin] (0, 0) grid (7, 9);
\draw [fill] (2, 8) rectangle (3, 9);
\draw [fill] (4, 8) rectangle (5, 9);
\draw [fill] (3, 7) rectangle (4, 8);
\draw [draw=gray, fill=black] (3, 8) rectangle (4, 9);

\node at (3.5, 0.5) {\textbf{D}};
\node at (2.5, 0.5) {\textbf{T}};
\node at (4.5, 0.5) {\textbf{T}};
\node at (3.5, 1.5) {\textbf{T}};

\node at (1.5, 3.5) {\textbf{W}};
\node at (2.5, 3.5) {\textbf{W}};
\node at (1.5, 4.5) {\textbf{W}};
\node at (2.5, 4.5) {\textbf{W}};
\node at (1.5, 5.5) {\textbf{W}};
\node at (2.5, 5.5) {\textbf{W}};

\node at (4.5, 3.5) {\textbf{W}};
\node at (5.5, 3.5) {\textbf{W}};
\node at (4.5, 4.5) {\textbf{W}};
\node at (5.5, 4.5) {\textbf{W}};
\node at (4.5, 5.5) {\textbf{W}};
\node at (5.5, 5.5) {\textbf{W}};

\node [text=white] at (3.5, 8.5) {\textbf{D}};
\node [text=white] at (2.5, 8.5) {\textbf{T}};
\node [text=white] at (4.5, 8.5) {\textbf{T}};
\node [text=white] at (3.5, 7.5) {\textbf{T}};

\node [circle, draw, inner sep=1pt] at (6.5, 4.5) {\textbf{7}};

\node [circle, fill, text=white, inner sep=1pt] at (3.5, 6.5) {\textbf{8}};

\end{tikzpicture}
      \caption{Lion vs. Elephant}
      \label{fig:lvse}
    \end{subfigure}
    \caption{Various endgames with two pieces.}
  \end{center}
\end{figure}

Figure~\ref{fig:decisiontrees}a shows the decision tree for each 
partition in which the two players have a piece of the same strength that
cannot make leaps, e.g., white elephant vs. black elephant. It has 
been observed by~\citeaby{Rijn2013} that these games do not end in a draw. 
Instead, there is a notion of \emph{parity}, which determines the outcome of
the game. This is illustrated in Figure~\ref{fig:evse} (White to move). 
Although White moves first and therefore can potentially reach the black
den first, it cannot take the path between the rivers, since it will
be captured by Black (recall that pieces of the same strength can 
capture each other). The white player cannot defend its own den due
to the parity, and the black player has a straightforward win in 
$6$~moves. Since tigers and lions can leap over the rivers, covering 
$3$~squares in one move, they can reverse the parity. Still, no draws occur, 
but the decision tree is more complex. Figure~\ref{fig:tvst} illustrates
this. Although Figure~\ref{fig:decisiontrees}a classifies this situation
as a win for Black, this is actually a win in $10$~for White. Playing
a sequence\footnote{Similar to the algebraic notation in Chess,
the characters denote which piece moves and to which square. An \texttt{x}
indicates a capture.} of:
\texttt{$1.\!$~T$a6$ t$a8$},
\texttt{ $2.\!$~T$d6$ t$a7$},
\texttt{ $3.\!$~T$d5$ t$b7$},
\texttt{ $4.\!$~T$d4$ t$b3$},
\texttt{ $5.\!$~T$d3$ t$a3$}
(\ldots
\texttt{t$b2$},
\texttt{ $6.\!$~T$c3$ t$b1$},
\texttt{ $7.\!$~T$c2$ t$a1$},
\texttt{ $8.\!$~T$b2$ t$b1$},
\texttt{ $9.\!$~Tx$b1$}),
\texttt{ $6.\!$~T$c3$ t$a2$},
\texttt{ $7.\!$~T$c7$ t$a1$},
\texttt{ $8.\!$~T$d7$ t$b1$},
\texttt{ $9.\!$~T$d8$ t$c1$},
\texttt{ $10.\!$~T$d9$},
White is able to reach the black den just before Black can reach the white den.

Figure~\ref{fig:decisiontrees}b shows the decision tree for partitions 
in which the black player has a stronger piece. This tree does not apply
to rats, tigers, and lions. % TODO check the sentence. 
When Black can reach the opposing den first, it is certainly a
win for Black, except for the situations in which the black piece
is trapped and the white player starts next to it. In the situations in which 
White is closer to the den, but cannot move unopposed to it, the situation is 
more complex. The strategy for White is to cross
the board, and threaten to enter the black den. If (s)he can reach rank~$7$, then depending on
(1)~parity, (2)~distance between the pieces, and (3)~distance to the den for Black, 
the white player might be able to force a draw. Figure~\ref{fig:cvsd} illustrates an exceptional 
example in which White can force a draw. White cannot 
cross using the shortest route, but it can threaten the black den by a
longer route. Black is forced to defend its den. 

The situation gets more complex when lions or tigers are involved. 
Figure~\ref{fig:decisiontrees}c shows the decision tree for partitions
with a white tiger or lion versus a black elephant. 
Tigers or lions can reverse the parity making it less important. 
Like in the previous tree, if White is closest to the black den, but
not unopposed, Black's role is to defend, forcing a draw. 
When White is in the $\mathrm{mid}$ sector, in some situations Black 
can force White out of the way, enabling him to reach the white den. 
In contrast to the other trees, this tree does not perfectly describe
all configurations; $16$ are misclassified. Figure~\ref{fig:lvse} illustrates
one of these misclassifications. Although Black is in the $\mathrm{top}$
sector, (s)he is able to force White out of the way as described before.

Using a set of reasonable game features (see Table~\ref{tab:features}),
we can construct perfect (i.e., no misclassifications) decision trees for 
all partitions of two piece endgame tablebases. Moreover, we can 
construct a tree describing the whole two-piece endgame tablebase 
in a straightforward way. In some situations it is preferable to have
a simpler tree, at the expense of a few misclassifications, such as in 
Figure~\ref{fig:decisiontrees}c.

\section{Conclusions}
We created a playing engine and constructed endgame tablebases for 
up to four pieces for the game Dou Shou Qi. Both can be used to 
gain novel insights into the intricacies of the game. It can also be considered as 
a first step towards theoretically solving Dou Shou Qi in the 
same way~\citeaby{Schaeffer07} solved Checkers.
We have represented the two-piece endgame tablebases as decision 
trees, using a set of reasonable game features. From these trees 
some interesting insights have been gained, most notably the 
\emph{importance of parity} and the \emph{absence of draws} in equal-material 
endgames with two pieces only.
Expanding the tablebase to more than four pieces is considered to be future 
work, as is the construction of decision trees for endgames with
more than two pieces. We are convinced that we then find even more interesting patterns.

\bibliography{note}

\end{document}